%% file: neurips_2025.tex
\documentclass{article}

\usepackage[preprint]{neurips_2025}

\usepackage[utf8]{inputenc} 
\usepackage[T1]{fontenc}    
\usepackage{hyperref}       
\usepackage{url}            
\usepackage{booktabs}       
\usepackage{amsfonts}       
\usepackage{nicefrac}       
\usepackage{microtype}      
\usepackage[table]{xcolor}         
\usepackage{graphicx}
\usepackage{amsmath}
\usepackage{caption}
\usepackage{subcaption}
\usepackage{multirow}
\usepackage{marvosym}

\title{P3P: Pseudo-3D Pre-training \\ for Scaling 3D Voxel-based Masked Autoencoders}

\author{
\centerline{
Xuechao Chen\textsuperscript{\rm 1},
Ying Chen\textsuperscript{\rm 2},
Jialin Li\textsuperscript{\rm 2},
Qiang Nie\textsuperscript{\rm 4 \rm 2},
Hanqiu Deng\textsuperscript{\rm 2},
Yong Liu\textsuperscript{\rm 2},
Qixing Huang\textsuperscript{\rm 3},
Yang Li\textsuperscript{\rm 1}\thanks{Corresponding author} \quad
} \\
\centerline{
\textsuperscript{\rm 1}SIGS, Tsinghua University \quad
\textsuperscript{\rm 2}Youtu Lab, Tencent \quad
\textsuperscript{\rm 3}The University of Texas at Austin \quad
} \\
\centerline{
\textsuperscript{\rm 4}Hong Kong University of Science and Technology (Guangzhou)
}
}

\begin{document}

\maketitle

\input{sections/00_abstract}
\input{sections/01_introduction}
\input{sections/03_method}
\input{sections/04_experiments}

\input{sections/02_related_works}
\input{sections/05_conclusions}

{
\small

\bibliographystyle{plain}
\bibliography{reference}
}






\end{document}

%% file: sections/00_abstract.tex
\begin{abstract}
3D pre-training is crucial to 3D perception tasks. Nevertheless, limited by the difficulties in collecting clean and complete 3D data, 3D pre-training has persistently faced data scaling challenges. In this work, we introduce a novel self-supervised pre-training framework that incorporates millions of images into 3D pre-training corpora by leveraging a large depth estimation model. New pre-training corpora encounter new challenges in representation ability and embedding efficiency of models. Previous pre-training methods rely on farthest point sampling and k-nearest neighbors to embed a fixed number of 3D tokens. However, these approaches prove inadequate when it comes to embedding millions of samples that feature a diverse range of point numbers, spanning from 1,000 to 100,000. In contrast, we propose a tokenizer with linear-time complexity, which enables the efficient embedding of a flexible number of tokens. Accordingly, a new 3D reconstruction target is proposed to cooperate with our 3D tokenizer. Our method achieves state-of-the-art performance in 3D classification, few-shot learning, and 3D segmentation.
Code is available at \url{https://github.com/XuechaoChen/P3P-MAE}.
\end{abstract}

%% file: sections/01_introduction.tex
\section{Introduction}

3D perception using 3D sensors like depth cameras and LiDAR devices is a fundamental task for interpreting and interacting with the physical world in fields such as robotics and augmented reality.
Similar to 2D and language, 3D pre-training can endow 3D models with more powerful perception performance, such as 3D classification and 3D segmentation.
Current 3D pre-training approaches~\cite{yu2022point, pang2022masked, zhang2022point, chen2024pointgpt, yan2023multi} usually require clean and complete 3D data to pre-train a 3D model, which is sampled from human-made CAD models~\cite{chang2015shapenet} or reconstructed from multi-view RGB/RGB-D scans~\cite{dai2017scannet} .
These approaches, however, are costly, as clean and complete 3D data requires human effort for denoising and manual correction. 
This limitation results in the lack of size and diversity in 3D pre-training data.
Concretely, the total number of all 3D samples over the Internet is only up to tens of millions~\cite {objaverseXL}.
Nevertheless, one can easily obtain billions of images from the Internet~\cite{ILSVRC15, schuhmann2022laion}, and the growth rate of images is much higher than that of 3D data.

\begin{figure*}
    \centering
    \includegraphics[width=1.0\textwidth]{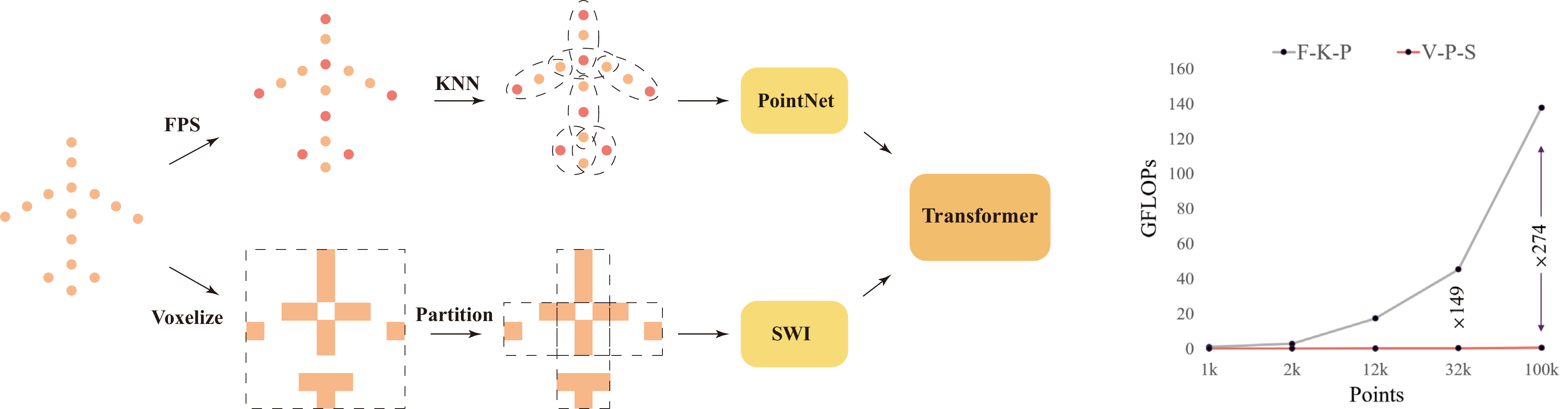}
    \caption{A comparison of the previous 3D tokenizer (top branch) with our 3D tokenizer (bottom branch). The right chart shows the giga floating-point operations (GFLOPs) needed in the previous tokenizer (F-K-P) and ours (V-P-S). Our 3D tokenizer requires many fewer operations than the previous tokenizer when embedding the same point cloud. }
    \label{fig:embed}
\end{figure*}

In this work, we alleviate the data-scale bottleneck in 3D pre-training by offline distilling the 3D geometric knowledge from a depth estimation model to a 3D model. 
We consider the depth estimation model as a teacher model predicting depth for millions of images in the ImageNet-1K~\cite{ILSVRC15} dataset.
Then we build the pseudo-3D pre-training corpora by lifting the images to 3D space, treating every pixel in each image as a 3D point. 
The point number equals the pixel number, which ranges from $1,000$ to $100,000$.

Since we create millions of 3D pre-training samples, and each sample contains a varying number of points (from $1,000$ to $100,000$), another challenge emerges in 3D token embedding.
As shown in the top branch of Fig.~\ref{fig:embed}, previous pre-training methods~\cite{yu2022point, pang2022masked, zhang2022point, chen2024pointgpt, yan2023multi} employ clustering using the farthest point sampling (FPS) and the k nearest neighbors (KNN), requiring quadratic time complexity.
Their tokenizer has 3 steps:
1) Select a fixed number of center points by the FPS algorithm.
2) Group the k nearest neighbors of centers into a fixed number of 3D patches.
3) Embed the 3D patches into 3D tokens by a PointNet~\cite{qi2017pointnet}.
When they pre-train their 3D models on the ShapeNet~\cite{chang2015shapenet} dataset, this tokenizer is acceptable because every 3D sample of ShapeNet only contains 1 thousand points, and the total number of samples is around 50 thousand.
However, in our pre-training data, the number of points of each sample varies from $1,000$ to $100,000$, and the total number of samples reaches 1.28 million.
Therefore, we need a more flexible and efficient 3D tokenizer to better adapt our pre-training data.
Similar to the image tokenizer in Vision Transformers~\cite{dosovitskiy2020image}, we introduce a 3D sparse tokenizer as illustrated in the bottom branch of Fig.~\ref{fig:embed}.
We have 3 steps: 
1) Voxelize the point cloud to voxels.
2) Partition the voxels into a flexible number of 3D patches.
3) Embed the 3D patches into 3D tokens by our proposed Sparse Weight Indexing.
To perform attention mechanism on the flexible number of tokens, we further add an attention mask in the Transformers.
In the right chart of Fig.~\ref{fig:embed}, we compare the giga floating-point operations (GFLOPs) between the previous tokenizer (F-K-P) and our tokenizer (V-P-S), which directly shows the time complexity of the two tokenizers.

As the 3D tokenizer changes, the 3D reconstruction target changes accordingly.
We design a new 3D reconstruction target for the pseudo-3D data, enabling the pre-trained model to capture the geometry, color, and occupancy distribution.
Extensive experiments demonstrate the efficacy and efficiency of our proposed methods.
We also achieve state-of-the-art performance on 3D classification, few-shot classification, and 3D segmentation among pre-training methods.

In summary, the main contributions of our work are as listed as follows:
\begin{itemize}
\item  We propose a novel self-supervised pre-training framework called P3P, which successfully distills the geometry knowledge of a teacher depth model and introduces natural color and texture distribution of millions of images from $1,000$ categories to 3D pre-training.
\item We introduce a voxel-based 3D tokenizer adapting to the new pre-training data, owning higher efficiency and flexible representation ability than the previous tokenizer.
\item We design a new 3D reconstruction target that adapts with our 3D tokenizer, enabling the 3D pre-trained model to better capture the geometry, color, and occupancy distribution of the pre-training data, which enhances the performance on downstream perception tasks.
\end{itemize}

%% file: sections/03_method.tex
\section{Approach}

In this section, we first briefly introduce the self-supervised pre-training method of Masked Autoencoders (MAE) and its 3D extension work, Point-MAE.
Then we detail our approach, including pre-training data creation, embedding, MAE pre-training, and reconstruction target.

\paragraph{Preliminaries.}
Masked Autoencoders (MAE)~\cite{he2022masked} method formulates the image self-supervised pre-training as a token masking and reconstruction problem.
At first, an image of the resolution $224\times 224$  is partitioned into $14\times 14=196$ patches with a patch size of $16\times 16$.
Each image patch of the resolution $16\times 16$ is embedded by multiplying with $16\times 16$ corresponding learnable weights.
Second, the embedded $196$ tokens of an image are randomly masked, and only a small part of the visible tokens are fed into a transformer encoder.
Third, the encoded visible tokens are fed to a transformer decoder with the masked empty tokens together.
The decoder is targeted at reconstructing all pixels in the masked tokens.
Mean squared error (MSE) is used to supervise the pre-training process, recovering the masked RGB pixels.

Later, Point-MAE extends the MAE pre-training from images to point clouds.
The main difference between Point-MAE and MAE lies in the token embedding and the reconstruction target.
Point-MAE embeds the point cloud as shown in Fig.~\ref{fig:embed} (top branch).
First, they sample a fixed number of center points by FPS. Second, the k nearest neighbors of each center point are grouped. Third, a fixed number of patches are fed into a PointNet to embed the 3D tokens. 
This tokenizer incurs quadratic time complexity, and the number of tokens is fixed, which harms the efficiency and representation ability in both pre-training and downstream tasks.
Since their pre-training dataset, ShapeNet, only has geometry features $x, y, z$, they change the MSE supervision to Chamfer Distance~\cite{fan2017point}.

In this work, we follow some validated settings of MAE and Point-MAE:
1) MAE uses MSE loss to supervise color reconstruction.
2) MAE employs $16\times 16$ weights and multiplies them with image patches of resolution $16\times 16$ to embed tokens.
3) Point-MAE uses Chamfer Distance to supervise geometry reconstruction.
4) Point-MAE utilizes linear layers with GELU~\cite{hendrycks2016gaussian} to obtain the positional embeddings of each 3D token.
The following subsections are the details of our proposed methods.

\begin{figure*}
    \centering
    \includegraphics[width=1.0\textwidth]{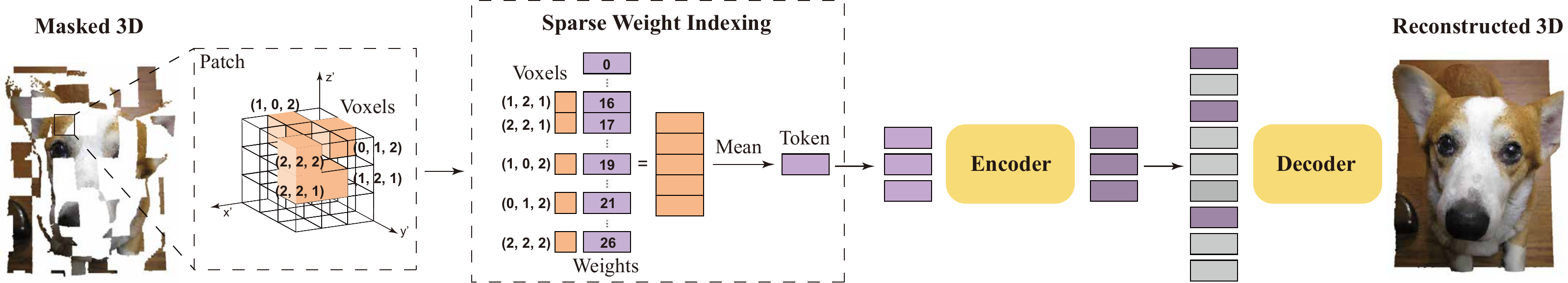}
    \caption{Overall pipeline of our 3D pre-training approach.}
    \label{fig:all}
\end{figure*}

\paragraph{Pre-training data creation.}
\label{lift_2d_to_3d}
Today there are zillions of images on the Internet.
One can easily acquire millions or billions of images~\cite{ILSVRC15, schuhmann2022laion} captured diverse objects or scenes from the real world.
To leverage any 2D images to pre-train 3D models, we choose the ImageNet-1K (IN1K)~\cite{ILSVRC15} without any depth annotations as our basic dataset.
Advantages of lifting 2D images to 3D include the much larger data size, better diversity, and natural color and texture distribution compared with point clouds collected by depth cameras, LiDAR, or sampled from 3D CAD models.

We employ an off-the-shelf large depth estimation model, Depth Anything V2 (large) ~\cite{depth_anything_v2}, which is trained on RGB-D scans and RGB images.
We denote it as $f$ for depth estimation, lifting 2D images into 3D space.
The self-supervised learning performed on the lifted pseudo-3D data can be regarded as offline distillation, introducing abundant 3D geometry, natural color, and texture information to our 3D model.
Given an image $I$, we obtain a lifted pseudo-3D point cloud 
$P=\phi(I, f(I))=\{ p_i=[x_i, y_i, z_i, r_i, g_i, b_i] | i=1,2,3,...,N \},$
where $x_i$, $y_i$, $z_i$ are the normalized continuous coordinates and $r_i$, $g_i$, $b_i$ are the normalized colors of each point $p_i$ corresponding to each pixel in image $I$.
$\phi$ maps the 2D coordinates of the image $I$ to 3D according to depth $f(I)$.
For a 3D coordinate system in robotics, one of the conventions is to portray the xy-plane horizontally, with the z-axis added to represent the height (positive up).
In this paper, we keep this convention.
Since most of the images in ImageNet are taken from horizontal views, we assign the predicted depth of the image to the y-axis.
The height of the image is aligned with the z-axis, and the width of the image is aligned with the x-axis.
Note that lifting occurs along the y-direction, making the generated pseudo-3D point cloud only have one viewing angle.
Therefore, random rotation along the z-axis is applied to make the generated 3D point clouds obey the normal distribution.

\paragraph{Embedding.}
\label{sec:embedding}
As mentioned above, each generated pseudo-3D sample has the same number of points as the image pixels, which ranges from $1,000$ to $100, 000$.
Previous 3D tokenizer used by Point-BERT~\cite{yu2022point}, Point-MAE~\cite{pang2022masked}, Point-M2AE~\cite{zhang2022point}, and PointGPT~\cite{chen2024pointgpt}, employs low-efficient farthest point sampling and k nearest neighbor to build a fixed number of 3D patches.
The tokenizer is not adaptive to large and varying point clouds, mainly due to its low efficiency.
The most feasible way to use the previous tokenizer on our data is to down-sample the original point cloud to the minimum number of points over all samples, which is $1,000$ points.
We run the original Point-MAE on our down-sampled data as a comparison baseline in Sec.~\ref{exp:classification}.

In this paper, we follow the token embedding implemented in MAE and Vision Transformers~\cite{dosovitskiy2020image} and extend it to a 3D sparse space based on the voxel representation.
Our tokenizer is capable of embedding point clouds with different numbers of points while maintaining high efficiency.
To enable the following transformer encoder and decoder to adapt to different numbers of tokens, we use the attention mask~\cite{vaswani2017attention}, which is different from previous pre-training methods.

Given a point cloud $P$, the voxelization $\psi$ processes the continuous coordinates $x$, $y$, $z$ and produces the discrete coordinates $m$, $n$, $q$ for voxels:
\begin{equation}
    (m, n, q) = (\lfloor{x/s} \rfloor, \lfloor{y/s} \rfloor, \lfloor{z/s} \rfloor),
\label{eq:voxelization}
\end{equation}
where $\lfloor*\rfloor$ denotes the floor function and $s$ is the basic voxel size, much less than the value of $x$, $y$, $z$. 
Since one voxel may contain many points, we simply choose the point with the maximum value of features to represent the voxel.
The voxel set $V$ representing the whole point cloud can be written as:
$V = \psi(P) = \{v_{m_i, n_i, q_i}=[x_i, y_i, z_i, r_i, g_i, b_i] | i=1,2,3,...,M, M\leq N\}.$
The subscript $m_i, n_i, q_i$ represents the discrete coordinates of the voxel $v_{m_i, n_i, q_i}$ calculated by Eq.~\ref{eq:voxelization}, and the value vector $[x_i, y_i, z_i, r_i, g_i, b_i]$ represents the voxel features of the voxel $v_{m_i, n_i, q_i}$. Note that the voxelization has a time complexity of $O(N)$.

Given range limits, the discretized 3D space is divided into multiple 3D patches.
We take the 3D patch $V_{a,a,0}\subset V$ that satisfies $\forall v_{m_i, n_i, q_i} \in V_{a,a,0},$ $ a \leq m_i, n_i<2a, 0 \leq q_i <a$, where $a$ is a positive integer representing the patch size, as an example.
In transformers, each token owns a corresponding positional embedding representing its global position.
For 3D point clouds, we can directly use the minimum corner's 3D coordinates to represent the position of each 3D patch.
Thus, the position of patch $V_{a,a,0}$ is $(a, a, 0)$.
Following Point-MAE, we employ linear layers with GELU~\cite{hendrycks2016gaussian} for positional embedding, which maps the coordinates to a high-dimensional feature space.
The positional embedding for the encoder and decoder is learned independently.

Point-MAE employs PointNet to extract KNN graph knowledge within a KNN point patch.
Inspired by their work, we build a voxel graph and save the graph knowledge into voxel features for later embedding.
We first calculate the graph center of patch $V_{a,a,0}$ as,
\begin{equation}
    (\overline{x}, \overline{y}, \overline{z}) = (\frac{1}{L}\sum_{i=1}^{L} x_i, \frac{1}{L}\sum_{i=1}^{L} y_i, \frac{1}{L}\sum_{i=1}^{L} z_i),
\end{equation}
where $L$ is the number of all voxels in patch $V_{a,a,0}$. 
Then we calculate the graph edge for each graph node (voxel) in patch $V_{a,a,0}$:
\begin{equation}
    (x_i', y_i', z_i') = (x_i-\overline{x}, y_i-\overline{y}, z_i-\overline{z}).
\end{equation}
Finally, the 3D patch $V_{a,a,0}'$ with graph knowledge is denoted as 
\begin{equation}
    V_{a,a,0}' = \{v_{m_i, n_i, q_i}' = [x_i, y_i, z_i, r_i, g_i, b_i, x_i', y_i', z_i', \overline{x}, \overline{y}, \overline{z}] | i=1,2,3,...,L, L\leq a^3<<M\}.
\end{equation}


The 3D patch $V_{a,a,0}'$ is now ready for token embedding.
We employ $a^3$ trainable weights to embed $V_{a,a,0}'$, which follows the 2D token embedding implemented by vision transformers~\cite{dosovitskiy2020image} and MAE~\cite{he2022masked} but differs in data input.
For 2D image patches, the data is dense and thus can use dense calculation, i.e., 2D convolution with a large kernel size $a\times a$ and stride $a$.
In contrast, 3D point clouds and their discrete voxels are sparse data.
Therefore, we propose Sparse Weight Indexing (SWI) to index the $a^3$ weights to $L$ sparse voxels and multiply them.
As shown in Fig.~\ref{fig:all}, for $v_{m_i, n_i, q_i}' \in V_{a,a,0}'$, we calculate the index $d_i$ of the corresponding weight $w_{d_i}$ as
\begin{equation}
\label{eq:indexing}
    d_i  = (m_i\%a) + (n_i\%a) a + (q_i\%a) a^2,
\end{equation}
where $(*\%a)$ calculates the remainder of $*$ divided by patch size $a$, and $d_i=0,1,2,...,a^3-1$.
Thus, we only need $a^3$ shared trainable weights to embed the 3D tokens for all 3D patches.
Each weight $w_{d_i} \in \mathbb{R}^{12\times C}$, where $12$ is the input feature dimension and $C$ is the embedding dimension.
We embed the token $T_{a,a,0}$ as,
\begin{equation}
\label{eq:embedding}
    T_{a,a,0} = \frac{1}{L} \sum_{i=1}^L v_{m_i, n_i, q_i}' w_{d_i}.
\end{equation}

With the help of hashing algorithms based on discrete voxel coordinates, regarding the 3 discrete coordinates as hashing keys (supported by the pytorch\_scatter library~\cite{pytorch_scatter}), the time complexity of embedding all tokens for all patches is $O(M)$.
In addition, batch operation is also supported by considering the batch number as the fourth hashing key.

\paragraph{Masked Autoencoders pre-training.}
As demonstrated in Fig.~\ref{fig:all}, the pseudo-3D sample is first randomly masked.
Second, the visible 3D patches are embedded into 3D tokens through the Sparse Weight Indexing method.
Third, the visible tokens are fed into a transformer encoder.
Finally, the encoded visible tokens and the masked empty tokens are fed into a transformer decoder to reconstruct every voxel within each masked patch.

\paragraph{Reconstruction target.}
To cooperate with the proposed 3D tokenizer, we design a new 3D reconstruction target in this subsection.
We follow the validated settings of MAE and Point-MAE.
MAE validated that the mean square error (MSE) is useful for supervising relative values like color reconstruction.
Point-MAE validated that the Chamfer Distance is effective on absolute values reconstruction, i.e., global geometry $x, y, z$.
We successfully combine them and add a new occupancy loss.
To reconstruct the relative values of every voxel in the masked patch $V_*'$, our MSE is defined as
\begin{align}
    & l_{MSE} = \frac{1}{|V_*'|} \sum_{i=1}^{|V_*'|} ||\mathbf{e_i}-\hat{\mathbf{e_i}}||_2^2,
\end{align}
where $\mathbf{e} = [r, g, b , x', y', z' , \overline{x}, \overline{y}, \overline{z}]$.
$\mathbf{e}$ is the ground truth vector and $\hat{\mathbf{e}}$ is the reconstructed vector.
To reconstruct the absolute values of every voxel in the masked patch $V_*'$, our Chamfer Distance is defined as
\begin{align}
    l_{CD} = \frac{1}{|V_*'|} & \sum_{\mathbf{c}\in V_*'} \min_{\hat{\mathbf{c}}\in\hat{V_*'}} ||\mathbf{c}-\hat{\mathbf{c}}||_2^2 + \frac{1}{|\hat{V_*'}|} \sum_{\hat{\mathbf{c}}\in\hat{V_*'}} \min_{\mathbf{c}\in V_*'} ||\mathbf{c}-\hat{\mathbf{c}}||_2^2,
\end{align}
where $\mathbf{c} = [x, y, z]$.
$\mathbf{c}$ is the ground truth vector, $\hat{\mathbf{c}}$ is the reconstructed vector, and $\hat{V_*'}$ is the reconstructed patch.
Owing to the voxel representation, we can easily add a loss to make the pre-trained model aware of the 3D occupancy~\cite{peng2020convolutional}, which can be defined as
\begin{align}
    l_{OCC} = -\frac{1}{a^3}\sum_{i=1}^{a^3}[o_i \log(\hat{o_i}) +(1-o_i)\log(1-\hat{o_i})],
\end{align}
where $o_i,\hat{o_i}\in\{0, 1\}$ indicating whether the i-th 3D discrete position in a masked patch is occupied.
$o_i$ is the ground truth occupancy, and $\hat{o_i}$ is the predicted occupancy.
The overall reconstruction loss is defined as,
\begin{align}
    l = l_{MSE} + l_{CD} + l_{OCC}.
\end{align}
In Sec.~\ref{sec:loss design}, our ablation study shows the effectiveness of each part of our reconstruction loss.

%% file: sections/04_experiments.tex
\section{Experiment}

\subsection{Transfer Learning}

\paragraph{Models.}
1) Tokenizer: In our models, we fix the patch size $a$ to $16$, and the discrete space size is $224 \times 224 \times 224$.
Therefore, we have $16 \times 16 \times 16 = 4,096$ learnable weights mapping the input $12$ features to $384$ (Encoder S) or $768$ (Encoder B) high-dimensional space.
2) Encoder S: The pre-training transformer encoder contains $12$ transformer blocks and $384$ feature channels. 
3) Encoder B: The pre-training transformer encoder contains $12$ transformer blocks and $768$ feature channels. 
4) Decoder: The pre-training transformer decoder is the same as the decoder used in MAE, containing $8$ transformer blocks and $512$ feature channels.
To map the encoded features to $512$ dimensions, a linear layer is employed.
5) Teacher depth estimation model: We choose ViT-large of Depth Anything V2 as our teacher model, predicting depth for the images.

\paragraph{Pre-training data and settings.}
Our pre-training data (denoted as P3P-Lift in this section) contains $1.28$ million pseudo-3D samples lifted from the images in ImageNet-1K (1K means $1,000$ categories)~\cite{ILSVRC15} training set.
Each sample owns $1,000$ to $100,000$ points and $6$ features, $x, y, z, r, g, b$.
We show some examples in our appendix.
Augmentation, including scaling and translation, is used during pre-training, which follows Point-MAE.
Color normalization is the same as in MAE.

We pre-train our models on $16$ $V100 (32GB)$ GPUs for $120$ epochs.
We pre-train the Encoder-S for 3 days and the Encoder-B for 6 days.
We use AdamW as the optimizer with a starting learning rate of $5e-4$ and a cosine learning rate scheduler.
The total batch size is set to $256$, and we update the parameters every $4$ iterations.
The masking ratio is set to $60\%$ during pre-training.

\input{tables/table_classification}

\paragraph{3D classification on 3D objects scanned from the real world.}
\label{exp:classification}
ScanObjectNN contains $2902$ scanned objects from the real world that are categorized into $15$ classes.
The raw objects own a list of points with coordinates, normals, and colors.
Each object has points ranging from $2,000$ to over $100,000$.
ScanObjectNN has multiple splits of data.
We choose to use the main split following Point-MAE.
For the base version of the data, $2902$ objects can be fed with background or not, denoted as OBJ\_BG or OBJ\_ONLY.
The hardest version of the data is denoted as PB\_T50\_RS, which augments each object 5 times by translation, rotation, and scaling.
In this version, there are $14,510$ perturbed objects with backgrounds in total.

3D classification takes the 3D point cloud of an object as input and predicts a class label for the object.
The pre-trained tokenizer and transformer encoder are followed by a classification head, and the decoder is discarded, in the same setting as MAE and Point-MAE.
At the fine-tuning stage, we fine-tune all the parameters in the pre-trained tokenizer and transformer encoder.
We take the class token as the input for the classification head.

As shown in Tab.~\ref{table:classification}, we start from the original Point-MAE pre-trained on the ShapeNet dataset.
For fair comparison, we conduct 3 experiments:
1) We pre-train the same Encoder S with our P3P-MAE method on the ShapeNet dataset.
2) We pre-train the same Encoder S with the Point-MAE method on our P3P-Lift dataset.
3) We pre-train the same Encoder S with our P3P-MAE method on our P3P-Lift dataset.
The improvement is shown in blue, which shows 3 conclusions:
1) Our P3P-MAE method is more effective than the Point-MAE baseline on the ShapeNet dataset.
2) A different method, Point-MAE, is also effective when pre-training on our P3P-Lift data.
3) Our P3P-MAE method is more adaptive to our P3P-Lift data than Point-MAE.
It is worth noting that Point-MAE can only deal with a fixed number of input points.
Therefore, we down-sample our P3P-Lift dataset and the ScanObjectNN dataset to the minimum number of points over all samples, which is $1,000$ and $2,000$.
In contrast, our method can tackle a varying number of input points from $1,000$ to $100,000$ and from $2,000$ to $100,000$.

Moreover, we conduct an experiment for model scaling as shown in the bottom line of Tab.~\ref{table:classification}.
Our pre-trained Encoder-B outperforms our pre-trained Encoder-S distinctively while pre-trained with the same number of epochs.
Our P3P approach outperforms the previous methods and achieves the state-of-the-art.
Especially, with the same encoder, we outperform the PointGPT pre-trained on 7 datasets ($300,000$ samples from ModelNet40~\cite{wu20153d}, PartNet~\cite{mo2019partnet}, ShapeNet, S3DIS~\cite{armeni20163d}, ScanObjectNN, SUN RGB-D, and Semantic3D~\cite{hackel2017semantic3d}) and the MVNet pre-trained on their multi-view Objaverse~\cite{deitke2023objaverse} dataset ($4.8$ million samples).
The Encoder-L in Tab.~\ref{table:classification} is a deeper transformer encoder with 15 transformer blocks.
Note that for fair comparison, we compare the results of PointGPT without post-pre-training.

\input{tables/table_fewshot}

\input{tables/table_modelnet}

\paragraph{Few-shot classification on 3D objects scanned from the real world.}
Few-shot learning aims to train a model that generalizes with limited data. 
We evaluate our pre-trained Encoder-B on ScanObjectNN OBJ\_BG.
We strictly follow CrossPoint~\cite{afham2022crosspoint} and conduct experiments on conventional few-shot tasks (N-way K-shot), where the model is evaluated on N classes, and each class contains K samples.
As shown in Tab.~\ref{tab:fewshot}, we outperform CrossPoint and Point-BERT by a large margin and achieve new state-of-the-art performance, showing the strong generalization ability of our model.

\paragraph{3D classification on 3D CAD objects.}
ModelNet40~\cite{wu20153d} consists of $40$ different categories of 3D CAD models, including objects like airplanes, bathtubs, beds, chairs, and many others. In total, there are $12,311$ models, with $9,843$ models for training and $2,468$ for testing.
Each point cloud sampled from CAD models contains $8,000$ points.
At the fine-tuning stage, we fine-tune all the parameters in the pre-trained tokenizer and transformer encoder.
Since most of the listed methods implement the voting strategy, we also implement voting while fine-tuning. 
As shown in Tab.~\ref{table:modelnet_shapenet}, we outperform the previous pre-training methods and achieve the state-of-the-art with the same transformer encoder.

\paragraph{3D segmentation.}
ShapeNetPart is based on the larger ShapeNet dataset, which contains a vast collection of 3D CAD models. ShapeNetPart focuses on the part-segmentation aspect of 3D shapes.
The dataset consists of $16,881$ 3D models across $16$ object categories. Each 3D model is associated with its part-level annotations, which are crucial for training and evaluating part-segmentation algorithms.
We follow Point-MAE to fine-tune our pre-trained Encoder-S.
As shown in Tab.~\ref{table:modelnet_shapenet}, we outperform the previous pre-training methods and achieve the state-of-the-art in terms of the mIoU for all classes and the mIoU for all instances.

\input{tables/table_ablation}

\subsection{Ablation Study}
\label{sec:ablation}

In this section, we conduct experiments to find the best settings for our approach.
We explore the impact of masking ratio, augmentation, input features, loss design, and data scaling.
For different control experiments, we independently pre-train Encoder-S for 30 epochs on our P3P-Lift data.
We evaluate our model with fine-tuning (ft.) and linear probing (lin.) on the classification task of ScanObjectNN OBJ\_BG.

\paragraph{Masking ratio.}
Random masking is an effective masking strategy according to MAE and Point-MAE.
We pre-train our model with different masking ratios to find the best setting.
As shown in Tab.~\ref{tab:ablation_masking}, we find that the masking ratio of 60\% achieves the highest accuracy on the downstream classification task.
When the masking ratio increases, the accuracy drops faster than when it decreases.

\paragraph{Augmentation.}
Augmentation, including scaling and translation, is widely used in the previous pre-training methods listed in our tables.
We conduct an ablation study to find the best setting for our pre-training.
As a result, randomly scaling half of the space and translating half of the space achieves the best performance.

\paragraph{Graph representation.}
The original features of the input point cloud include coordinates $x, y, z$, and colors $r, g, b$.
We run an experiment pre-training only on the original 6 features.
As shown in Tab.~\ref{tab:ablation_features_loss}, our graph representation surpasses the original features by $4.63\%$ fine-tuning accuracy.

\paragraph{Loss design.}
\label{sec:loss design}
We conduct an ablation study on our 3 reconstruction losses, i.e., MSE, Chamfer Distance, and Occupancy.
As shown in Tab.~\ref{tab:ablation_features_loss}, our hybrid loss achieves $9.26\%$ higher fine-tuning accuracy than the MSE loss only.

\paragraph{Data scaling.}
We conduct data scaling experiments in Tab.~\ref{tab:ablation_datascale}, showing that the data scale $500K$ is a milestone.
Under the data scale $100K$, the pre-training cannot converge to a good result.
When the pseudo-3D samples come up to $500K$, the pre-training begins to show its powerful performance.

%% file: tables/table_classification.tex
\begin{table*}[t]
\footnotesize
\centering
\caption{Classification results (fine-tune all parameters end-to-end) on ScanObjectNN dataset. ``Samples'' shows the number of samples for pre-training. ``Points'' shows the number of input points. \label{table:classification}}
\resizebox{\linewidth}{!}{
\begin{tabular}{lcccccccc}
\toprule
&  & & & & \multicolumn{4}{c}{ScanObjectNN} \\ 
\cline{6-9} 
\multicolumn{1}{l}{\multirow{-2}{*}{Method}} & \multicolumn{1}{c}{\multirow{-2}{*}{Dataset}} & \multicolumn{1}{c}{\multirow{-2}{*}{Samples}} & \multicolumn{1}{c}{\multirow{-2}{*}{Points}} & \multicolumn{1}{l}{\multirow{-2}{*}{Encoder}} & Points & OBJ\_BG  & OBJ\_ONLY & PB\_T50\_RS \\ 
\midrule
\multicolumn{9}{c}{{\small $Supervised \ Learning$}} \\
\midrule
PointNet \cite{qi2017pointnet} & - & - & - & -  & 1K & 73.3    & 79.2  & 68.0 \\
PointNet$++$~\cite{qi2017pointnet++} &  - & - & - & -  & 1K & 82.3 & 84.3 & 77.9 \\
DGCNN \cite{wang2019dynamic} & - & - & - & - & 1K & 82.8   & 86.2  & 78.1   \\
MVTN \cite{hamdi2021mvtn} & - & - & - & - & - & 92.6 & 92.3 & 82.8 \\
\midrule
\multicolumn{9}{c}{{\small $Self\text{-}Supervised \ Learning$}} \\
\midrule
Point-BERT \cite{yu2022point} & ShapeNet & 50K & 1K & S  & 1K  &  87.4  & 88.1  & 83.1  \\
MaskPoint \cite{liu2022masked} & ShapeNet & 50K & 1K & S  & 2K  &  89.3   & 88.1 & 84.3  \\
Joint-MAE~\cite{guo2023joint} & ShapeNet & 50K & 2K & S  & 1K & 90.9 & 88.9 & 86.1 \\
PointGPT~\cite{chen2024pointgpt} & ShapeNet & 50K   & 1K & S  & 2K  &  91.6  &  90.0 & 86.9 \\
MVNet\cite{yan2023multi} & ShapeNet & 300K & - & S  & - &   91.4 & 89.7 & 86.7 \\
MVNet & Objaverse & 4.8M & - & S  & - & 91.5 & 90.1 & 87.8 \\
\rowcolor{gray!10} Point-MAE \cite{pang2022masked} & ShapeNet & 50K & 1K & S  & 2K &  90.0  & 88.2 & 85.2 \\
\rowcolor{gray!10} P3P-MAE & ShapeNet & 50K & 1K & S  &   2K &   90.7(\textcolor{cyan}{+0.7})  & 90.1(\textcolor{cyan}{+1.9})   & 85.8(\textcolor{cyan}{+0.6})  \\
\rowcolor{gray!20} Point-MAE & P3P-Lift & 1.2M & 1K & S & 2K &  92.3(\textcolor{cyan}{+2.3})  & 92.6(\textcolor{cyan}{+4.4})  & 89.1(\textcolor{cyan}{+3.9})  \\
\rowcolor{gray!20} P3P-MAE & P3P-Lift & 1.2M & 1$\sim$100K & S  & 2$\sim$100K &   \textbf{94.5}(\textcolor{cyan}{+4.5})  & \textbf{93.1}(\textcolor{cyan}{+4.9})  & \textbf{89.6}(\textcolor{cyan}{+4.4})  \\
\midrule
Point-M2AE \cite{zhang2022point} & ShapeNet  & 50K  & 2K & L  & 2K &   91.2 & 88.8 & 86.4 \\
I2P-MAE~\cite{zhang2023I2PMAE} & ShapeNet & 50K & 2K & L & 2K & 94.1 & 91.6 & 90.1 \\
PointGPT & 7 datasets & 300K & 1K & B  & 2K &   93.6 & 92.5 & 89.6 \\
MVNet & Objaverse & 4.8M & - & B  & - &   95.2 & 94.2 & 91.0 \\
\rowcolor{gray!20} P3P-MAE & P3P-Lift & 1.2M & 1$\sim$100K & B & 2$\sim$100K &    \textbf{95.5}   &  \textbf{94.9} & \textbf{91.3} \\
\bottomrule
\end{tabular}}
\vspace{-5pt}
\end{table*}

%% file: tables/table_fewshot.tex
\begin{table}[t]
\centering
\footnotesize
\caption{{Few-shot object classification results on ScanObjectNN.} We report mean and standard error over 10 runs. \label{tab:fewshot}}
\begin{tabular}{lcccc}
\bottomrule \multicolumn{1}{l}{\multirow{2}{*}{Pre-train Method}}  & 
\multicolumn{2}{c}{5-way} & \multicolumn{2}{c}{10-way} \\\cline{2-5}
& 10-shot & 20-shot & 10-shot & 20-shot \\ 
\midrule
Jigsaw \cite{jigsaw}& 65.2$\pm$3.8 & 72.2$\pm$2.7 & 45.6$\pm$3.1 & 48.2$\pm$2.8\\
cTree \cite{ctree} & 68.4$\pm$3.4 & 71.6$\pm$2.9 & 42.4$\pm$2.7 & 43.0$\pm$3.0\\
OcCo \cite{occo} & 72.4$\pm$1.4 & 77.2$\pm$1.4 & 57.0$\pm$1.3 & 61.6$\pm$1.2\\
CrossPoint \cite{afham2022crosspoint} & {74.8$\pm$1.5} &{79.0$\pm$1.2} & {62.9$\pm$1.7} & {73.9$\pm$2.2}\\
Point-BERT \cite{yu2022point} & {78.7$\pm$4.1} & {83.1$\pm$5.8} & {65.0$\pm$5.1} & {74.4$\pm$4.0}\\
\rowcolor{gray!20} P3P-MAE & \textbf{84.2$\pm$6.4} & \textbf{88.0$\pm$2.1} & \textbf{71.9$\pm$3.6} & \textbf{78.9$\pm$3.2} \\
\bottomrule
\end{tabular}
\vspace{-5pt}
\end{table}

%% file: tables/table_modelnet.tex
\begin{table*}[t]
\footnotesize
\centering
\caption{Classification results (fine-tune all parameters end-to-end) on ModelNet40 dataset and segmentation results on ShapeNetPart. We report mean intersection over union for all classes Cls.mIoU (\%) and all instances Inst.mIoU (\%) for part segmentation. \label{table:modelnet_shapenet}}
\begin{tabular}{lccccc}
\toprule
&  & \multicolumn{2}{c}{ModelNet40} & \multicolumn{2}{c}{ShapeNetPart Seg.}\\ 
\cline{3-6}
\multicolumn{1}{l}{\multirow{-2}{*}{Method}} & \multicolumn{1}{l}{\multirow{-2}{*}{Encoder}} & Points & ACC(\%) & Cls.mIoU & Inst.mIoU \\ 
\midrule
\multicolumn{6}{c}{{\small $Supervised \ Learning$}} \\
\midrule
PointNet \cite{qi2017pointnet} & - & 1K & 89.2    & 80.4 & 83.7 \\
PointNet$++$~\cite{qi2017pointnet++} &  - & 1K & 90.7 & 81.9 & 85.1\\
DGCNN \cite{wang2019dynamic} & - & 1K & 92.9   &  82.3 & 85.2  \\
MVTN \cite{hamdi2021mvtn} & -  & - & 93.8 & - & - \\
\midrule
\multicolumn{6}{c}{{\small $Self\text{-}Supervised \ Learning$}} \\
\midrule
Point-BERT \cite{yu2022point} & S  & 8K &    93.8  & 84.1 & 85.6\\
MaskPoint \cite{liu2022masked} & S  & 1K &     93.8 & 84.4 & 86.0\\
Point-MAE \cite{pang2022masked} & S  & 1K &    93.8  & 84.2 & 86.1\\
PointGPT~\cite{chen2024pointgpt}  & S  & 8K &    94.2 & 84.1 & {86.2}\\
\rowcolor{gray!20} P3P-MAE & S  & 8K & \textbf{94.4}  & \textbf{84.5} & \textbf{86.4} \\
\bottomrule
\end{tabular}
\end{table*}

%% file: tables/table_ablation.tex
\begin{table*}[t]
\centering
\footnotesize
\caption{Ablation study of our design choices. The evaluation metric is the fine-tuning (ft.) and linear probing (lin.) classification accuracy on the ScanObjectNN OBJ\_BG dataset.}
\begin{subtable}[t]{0.4\textwidth}
\centering  
    \caption{Masking Ratio\label{tab:ablation_masking}}
    \begin{tabular}{ccc}
        \toprule
        ratio & ft. & lin. \\ 
        \midrule
        15 & 86.87 & 60.89 \\
        30 & 88.33 & 66.55 \\
        45 & 88.79 & 67.28 \\
        \rowcolor{gray!20} 60 & \textbf{89.36} & \textbf{70.66} \\
        75 & 89.19 & 68.54 \\
        90 & 83.01 & 34.81 \\
        \bottomrule
    \end{tabular}
\end{subtable}%
\begin{subtable}[t]{0.6\textwidth}
\centering  
    \caption{Augmentation\label{tab:ablation_augmentation}}
    \begin{tabular}{lccc}
        \toprule
        Augmentation & ratio & ft. & lin.\\
        \midrule
        Baseline & - & 88.21 & 61.03 \\
        \multirow{3}{*}{+ Scaling} & 3/4 & 88.33 & 63.46 \\
        & 2/3 & 89.36 & 70.66 \\
        & 1/2 & 89.71 & 71.21 \\
        \rowcolor{gray!20} + Translation & 1/2 & \textbf{90.39} & \textbf{72.77} \\
        \bottomrule
    \end{tabular}
\end{subtable}
\vspace{4pt}
\\
\begin{subtable}[t]{\textwidth}
\centering 
    \caption{Graph Representation and Loss Design\label{tab:ablation_features_loss}} 
    \begin{tabular}{ccccccc}
        \toprule
        origin & graph & MSE & Chamfer & Occupancy & ft. & lin. \\
        \midrule
         & \checkmark & \checkmark & & & 80.10 & 36.70 \\
         & \checkmark & \checkmark & \checkmark & & 83.01 & 50.42 \\
        \rowcolor{gray!20} & \checkmark & \checkmark & \checkmark & \checkmark & \textbf{89.36} & \textbf{70.66} \\
        \checkmark & & \checkmark & \checkmark & \checkmark & 84.73 & 44.08 \\
       \bottomrule
    \end{tabular}
\end{subtable}
\vspace{4pt}
\\
\begin{subtable}[t]{\textwidth}
\centering  
    \caption{Data Scaling\label{tab:ablation_datascale}}
    \begin{tabular}{lccccccc}
        \toprule
        Data scale &  & 12K & 25K & 50K & 100K & 500K & 1.2M \\
        \midrule
        \multirow{2}{*}{P3P-Lift} & ft. & 30.70 & 35.50 & 30.36 & 29.15 & 81.98 & \textbf{89.36} \\
        & lin. & 25.38 & 26.24 & 25.38 & 23.49 & 57.46 & \textbf{70.66} \\
        \bottomrule
    \end{tabular}
\end{subtable}
\end{table*}

%% file: sections/02_related_works.tex
\section{Related Work}

\paragraph{Voxel-based representation.}
In the 3D domain, voxel-based methods~\cite{li2022unifying, chen2023svqnet, mao2021voxel, he2022voxel, li2023voxformer, wang2023dsvt, he2024scatterformer, yang2023pvt} are proven more efficient and flexible than point-based methods. 
Especially for the large point clouds with a varying number of points.
Sparse convolution~\cite{choy20194d, spconv2022, tang2022torchsparse} is widely used in these voxel-based methods.
However, in this work, it is hard to directly use sparse convolution because:
1) The correspondence between voxels and patches is needed to reconstruct each voxel in a masked 3D patch.
2) The attention mask is needed for the transformer encoder and decoder.
Therefore, we manually design our 3D voxel-based tokenizer, consisting of voxelization, partitioning, and Sparse Weight Indexing, and thus enhance its expandability and portability for future work.

\paragraph{3D self-supervised pre-training.}
PointContrast~\cite{xie2020pointcontrast} proposed an unsupervised pre-text task for 3D pre-training.
It learns to distinguish between positive and negative point pairs, where positive pairs come from the same object or scene, and negative pairs come from different objects or scenes.
Later, OcCo~\cite{occo} proposed an unsupervised point cloud pre-training, feeding the encoder with a partial point cloud and making the decoder predict the whole point cloud.
Point-BERT introduces a masked 3D pre-training and provides a solid codebase that enables followers to build their methods.
Based on Point-BERT, Point-MAE extends the MAE pre-training method to the 3D point cloud.
Point-M2AE~\cite{zhang2022point} proposes multi-scale masking pre-training, making the model learn hierarchical 3D features.
Following the 3D patch embedding used by Point-BERT, Point-MAE, and Point-M2AE, PointGPT~\cite{chen2024pointgpt} proposes an auto-regressively generative pre-training method for point cloud learning.
They both pre-train their transformers on the ShapeNet~\cite{chang2015shapenet} dataset, containing around 50,000 unique 3D models from 55 common object categories, which has limitations in pre-training data size and the 3D token embedding strategy.

\paragraph{2D \& 3D joint self-supervised pre-training.}
Some recent work focuses on jointly pre-training on 2D and 3D data.
Joint-MAE~\cite{guo2023joint} leverages the complementary information in 2D images and 3D point clouds to learn more robust and discriminative representations.
Multiview-MAE~\cite{chen2023point} is trained to reconstruct 3D point clouds from multiple 2D views and vice versa. This allows the model to capture the inherent correlations between 3D and 2D data.
SimIPU~\cite{li2022simipu} leverages the inherent spatial correlations between 2D images and 3D point clouds to learn spatial-aware visual representations.
PiMAE~\cite{chen2023pimae} and Inter-MAE~\cite{liu2023inter} are trained to reconstruct the original data from one modality (e.g., point cloud) using the information from the other modality (e.g., image). This allows the model to capture the inherent correlations between the two modalities.
These methods focus on distillation from 2D pre-trained models or learning from the correspondence between 2D and 3D.


%% file: sections/05_conclusions.tex
\section{Conclusion}

In this paper, we have reviewed recent progress in 3D pre-training, discussing key issues from a data-driven perspective.
We have proposed an approach based on MAE leveraging pseudo-3D data lifted from images, introducing a large diversity from 2D to 3D space.
To efficiently utilize the large-scale and varying data, we have proposed a novel 3D tokenizer and a corresponding 3D reconstruction target.
We have conducted downstream experiments demonstrating the effectiveness of our pre-training approach, reaching state-of-the-art performance on 3D classification, few-shot learning, and segmentation.
We have conducted different ablation studies to reveal the nature and enlighten future research.
We also have limitations mainly caused by the shortage of GPU resources.
We only have tens of GPUs to conduct pre-training on millions of samples.
In future work, we will find more GPUs and enlarge the data scale to billions of samples and pre-train a more powerful 3D foundation model for perception tasks.